\documentclass[conference]{IEEEtran}
\IEEEoverridecommandlockouts
\usepackage{cite}
\usepackage{amsmath,amssymb,amsfonts}
\usepackage{algorithmic}
\usepackage{graphicx}
\usepackage{textcomp}
\usepackage{xcolor}
\usepackage{subcaption}
\usepackage{enumitem}
\usepackage{multirow}
\usepackage[pagebackref=false,breaklinks=true,colorlinks,bookmarks=false]{hyperref}
\def\BibTeX{{\rm B\kern-.05em{\sc i\kern-.025em b}\kern-.08em
    T\kern-.1667em\lower.7ex\hbox{E}\kern-.125emX}}
\begin{document}

\title{Longitudinal Performance of Iris Recognition in Children: Time Intervals up to Six years\\

\makeatletter
\newcommand{\linebreakand}{%
  \end{@IEEEauthorhalign}
  \hfill\mbox{}\par
  \mbox{}\hfill\begin{@IEEEauthorhalign}
}
\makeatother

\thanks{This work is financially supported by the Center for Identification Recognition and Technology (CITeR) and National Science Foundation (NSF) Grant no. \# 1650503}
}

\author{\IEEEauthorblockN{1\textsuperscript{st} Priyanka Das}
\IEEEauthorblockA{\textit{Electrical and Computer Engineering} \\
\textit{Clarkson University}\\
New York, USA \\
prdas@clarkson.edu}
\and
\IEEEauthorblockN{2\textsuperscript{nd} Naveenkumar G Venkataswamy }
\IEEEauthorblockA{\textit{Electrical and Computer Engineering} \\
\textit{Clarkson University}\\
New York, USA \\
venkatng@clarkson.edu}

\and
\IEEEauthorblockN{3\textsuperscript{rd} Laura Holsopple }
\IEEEauthorblockA{\textit{Electrical and Computer Engineering} \\
\textit{Clarkson University}\\
New York, USA \\
lholsopp@clarkson.edu }

\and
\IEEEauthorblockN{4\textsuperscript{th} Masudul H Imtiaz}
\IEEEauthorblockA{\textit{Electrical and Computer Engineering} \\
\textit{Clarkson University}\\
New York, USA \\
mimtiaz@clarkson.edu }

\and

\IEEEauthorblockN{5\textsuperscript{th} Michael Schuckers}
\IEEEauthorblockA{\textit{Mathematics, Computer Science and Statistics} \\
\textit{St. Lawrence University}\\
New York, USA \\
schuckers@stlawu.edu }
\and 
\IEEEauthorblockN{6\textsuperscript{th} Stephanie Schuckers}
\IEEEauthorblockA{\textit{Electrical and Computer Engineering} \\
\textit{Clarkson University}\\
New York, USA \\
sschucke@clarkson.edu }

}

\maketitle

\begin{abstract}
The temporal stability of iris recognition performance is core to its success as a biometric modality. With the expanding horizon of applications for children, gaps in the knowledge base on the temporal stability of iris recognition performance in children have impacted decision-making during applications at the global scale. This report presents the most extensive analysis of longitudinal iris recognition performance in children with data from the same 230 children over 6.5 years between enrollment and query for ages 4 to 17 years. Assessment of match scores, statistical modelling of variability factors impacting match scores and in-depth assessment of the root causes of the false rejections concludes no impact on iris recognition performance due to aging.
\end{abstract}


\section{Introduction} The longitudinal stability, or permanence, is one of the core features in defining biometrics, the science behind establishing the identity of a person based on physical, chemical, or behavioural characteristics inherent to an individual. Systematic scientific studies to establish the longitudinal stability of different biometric modalities remain under-explored, more so for the demography of children. 

Of the most widely used biometric modalities (face, fingerprint, iris, voice), longitudinal studies of face recognition in children have been studied to a considerable degree. In the past decade, longitudinal face recognition has been studied \cite{deb2018longitudinal, ricanek2015review,best2016automatic,kamble2022machine,chandaliya2021longitudinal,guru2020children}, to establish the impact of changes in the facial structure in children. Longitudinal fingerprint \cite{jain2016fingerprint} and voice \cite{purnapatra2020longitudinal} recognition in children is studied, though inadequately. On the contrary, the iris has been assumed to have temporal stability with little scientific evidence. Between 2008 and 2016, multiple studies \cite{tome2008effects,hollingsworth2009pupil,baker2009empirical,fairhurst2011analysis,fenker2012analysis,rankin2012iris,baker2013template,ortiz2013linear,grother2013irex,ND_response_to_NIST,czajka2013template,grother2015irex}, reported on the longitudinal stability of iris with conflicting conclusions. However, none of the studies investigated the demography of children. With a growth factor involved in children, conclusions from studies on adults cannot be translated to children. With the expanding horizon of applications, this gap in the knowledge base became more prominent, impacting decision-making during applications. In 2010, India introduced Aadhar \cite{Aadhar}. It utilizes multi-biometric (face, fingerprint, iris) modality-based identification to de-duplicate identity on a national scale. The protocol includes two requirements that could possibly be a result of the knowledge gap in the biometric community: (a) Capture of biometric information is not done for children aged below 5 years; (b) Biometric information of children should be updated at the age of 15. The Nexus program \cite{Canada_Report} adopted by Canada for the trusted traveller program uses iris biometrics for identity verification. The program enrols children. Based on transaction logs of the NEXUS system as early as 2003, the Canadian Defence Research organization published a report on their observations and concluded that iris recognition is less reliable but still useful for the younger age group. The conclusions are drawn on the premises that (a) the size of the iris changes for children and is not stable until 6 to 8 years of age;
(b) subjects below age 14 had a higher percentage of `enrollment with only one eye. When the Nexus and the Aadhaar programs were launched in 2003 and 2010, respectively, there was limited literature on:\\
\textit{Can biometrics be collected from children? 
Is the available technology suitable for this demography? Do the biometric features change as a factor of aging in children? If biometric features do change, how does that impact biometric recognition performance? Is there an age where changes are no longer seen?} These open questions led to decisions that underscored the limits in scientific knowledge of children's demography for using iris for biometric recognition. 

In 2017, the first report on the feasibility of the available biometric technology in capturing and analyzing biometrics from children, including iris, was published\cite{basak2017multimodal}. Data from 106 children between 18 months and four years who participated in each of the two sessions, spaced by 4 to 6 months, were matched for verification with approximately 99.82\% (98.95 in the left iris) accuracy with a single gallery image and 100\% accuracy with multiple images. The report concluded that iris recognition yields the best performance though it is the most challenging modality to collect from children.
To partially address this knowledge gap, in 2016, our research group started investigating the longitudinal performance of 5 different biometric modalities for the demography of children, including iris. A systematic study was designed to understand the impact of aging in children on biometric recognition performance. The study is still ongoing. To the best of our knowledge, this study is the only one investigating longitudinal performance at this scale. A preliminary assessment of the data collected from 3 sessions over one year spaced by six months from 123 children was published in 2018 by Johnson et al. \cite{johnson2018longitudinal}. In 2021, Das et al. \cite{das2021iris} reported on the extended version of the same dataset collected over 3 years in seven sessions from 209 children. The report concluded the following:
\begin{itemize}[noitemsep, leftmargin=*]
\item [-] No impact of aging on longitudinal iris recognition performance over three years
\item [-]   Time gap between the mating pair of samples and the match score (MS) is negatively correlated which is \textit{statistically significant} but \textit{practically insignificant}. 

\item [-] Difference in dilation between mating pair of samples, $\Delta D$ and MS are negatively correlated and is both statistically and practically significant. The effect is higher by approximately 29 to 45 times than that of the time gap. However, it did not impact FNMR.

\item [-]  Aging may impact dilation as a relationship was observed between dilation and age.

\item [-] The report exposed gaps in the existing technology of iris capture, especially in the younger age group below five years, with the state-of-the-art cameras primarily designed for adults and the design of the camera contributed to 10\% failure to capture. 

\end{itemize}

There was a medical observation by Alder \cite{adler1965physiology} in 1965 on the varying dilation in children. In 2020, Das et al. \cite{das2020analysis} reported the first research on our dataset on the impact of pupil dilation on iris recognition performance for the age group of 4 to 14 years. The report concluded no impact on iris recognition performance due to change in dilation with age for the age group of 3 to 14 years over a time period of 3 years, with the caution that the quality of the images is maintained. Delta dilation impacts MS by 8.5\% on average. Delta dilation due to aging impacts MS by a small percentage of 8.5\%.  

The report presented in this paper is an extension of the report prepared by Das et al.\cite{das2021iris}. This paper reports on detailed research and analysis of the `permanence' characteristics of the iris of the same dataset, now containing 6.5 years of longitudinal data from the same children as a factor of aging. This report is based on the study of 230 individual subjects aged between 4 to 11 years at enrollment from 10 sessions over 78 months (6.5 years). The observations and conclusion would impact global applications with use cases involving children like Aadhar in India, border security, child trafficking, distribution of benefits, etc., and in the development of iris capture systems for this demography. This paper adds 3.5 additional years to the reports previously published in \cite{das2021iris} and \cite{das2020analysis}. Sharing our findings with the biometric community as our study progress will add to the critical knowledge base of the largely unexplored domain of longitudinal performance of iris recognition in children. Our contribution includes:
\begin{itemize}[noitemsep, leftmargin=*]
    
\item 	Assessment of longitudinal iris recognition performance over 78 months (6.5 years) from 13 time-frames spaced by 6 months from the same 230 children aged between 4 to 17 years collected from 10 sessions (4 missing sessions due to COVID-19).
\item  Statistical modelling of the impact of variability factors like enrollment age, dilation, and difference in dilation between the mating pair of samples, in addition to the time difference between the mating pair of samples on the MS using linear mixed-effects regression model (LMER). 
\item An in-depth analysis of root causes of false rejections

\end{itemize}

\section{Experimentation}
\subsection{Clarkson Iris Children Dataset} 
The analysis reported in this study is based on the dataset created by Clarkson University as part of the longitudinal study of biometric recognition performance in children. Iris is one of the five modalities collected as part of this project. Data is collected by the voluntary participation of children from the local Potsdam Elementary, Middle, and High Schools, with written consent from parents and age-tiered assent from children. Initial enrollment was offered to subjects aged 4 to 11 years. Subsequently, every year enrollments are offered to Pre-K students, predominantly aged between 4 to 5 years. This increases the number of subjects for the lower time frames ( 6 mo, 12 mo, etc.); however, the dataset will always have fewer samples and subjects for the long time gaps. Every Spring and Fall, spaced by approximately six months, our research team holds data collection sessions in the school. The same subjects are tracked over time. However, there are gaps in participation from individual subjects due to factors such as being absent from a particular session, refusing to participate in a session, or moving away from the local school. Ideally, data is collected every six months. However, the COVID-19 pandemic impacted data collection leading to 4 missing sessions between Spring 2019 and Fall 2021. Data collection resumed in the Spring of 2022. The lost session leaves a hole in our analysis of 3 of the 13-time frames. However, from a longitudinal viewpoint, the dataset contains a statistically significant number of subjects and data to make reliable conclusions over 6.5 years. The dataset contains 11223 samples from 230 individuals over 6.5 years from 10 sessions. 

 Iris Guard IG-AD100 Dual Iris Camera \cite{IrisGuard} is used for data capture. Iris data is collected under a semi-controlled environment with conscious steps taken to maintain the quality of the data and mitigate noise, like drawing blinds in the collection room to eliminate NIR-light interference from sunlight and allowing subjects enough time in the room to accommodate dilation to room lighting. More detailed notes on the data collection procedure are available in our earlier report \cite{das2021iris}.  
\subsection{Performance Evaluation and Metrics}
For evaluation of longitudinal performance, the first session the subject participated in the study is considered as the enrollment session. All subsequent sessions in which the subject participated are considered probe sessions for different time-frames (6 months to 78 months, spaced by six months). Multiple samples were collected per subject per session. All match scores and quality metrics are computed using VeriEye software version 12.4 \cite{VerieyeSDK}. All computations in VeriEye follow ISO/IEC 29794-2 \cite{iris_standard_report}. The analysis is performed in MATLAB and R. The longitudinal performance is evaluated following the false non-match rate (FNMR) metric, which has an operational impact. Additionally, Linear Mixed Effects Regression Model (LMER) is adapted to understand the impacts of factors like enrollment age (EA), probe image dilation (PD), the difference in dilation between mated pairs of samples ($\Delta D$) and the time difference between the capture of enrollment and probe sample (TF) on the variance in match score (MS). The metrics and the model covariates used in our analysis are defined below. 

\begin{enumerate}[noitemsep, leftmargin=*]
\item \textbf{False Non-match Rate}
   is the fraction of mated
(from the same individual) comparisons that fail to match. In this study, FNMR is computed at an MS
threshold of 36, equivalent to a False Match Rate (FMR) of 0.1\% following VeriEye calibration\cite{VerieyeSDK} 

\item \textbf{Dilation (D)}\label{Dil}
    or pupil dilation is a dimensionless quantity measuring the degree to which the pupil is dilated or constricted, measured as a ratio of pupil and iris radius. The measure follows ISO/IEC 29794-6 \cite{iris_standard_report}

    \item \textbf{Delta Dilation ($\Delta D$)} is the difference in pupil dilation between a mated pair of iris images. The measure follows NIST work in \cite{grother2013irex} as below.
        \begin{equation}
        \scriptsize
           \mathit{\mathbf{ Delta\:Dilation (\Delta D) = 1 - \frac{1-\frac{D1}{100}}{1-\frac{D2}{100}}}}
        \end{equation}
    considering, D1 $\geq$ D2, where, D1 and D2 are the pupil dilation of the first and the second iris images.
    \item \textbf{Enrollment Age (EA)} is 
age in \textit{years} at which the subject first participated in the study
     \item \textbf{Time Frame (TF)} is the 
    time elapsed between the capture of enrollment and probe sample measured in \textit{months}. 
\end{enumerate}

\section {Results}

This section reports on the evaluation of 11223 samples from 460 unique irides contributing to 30535 mated pairs of samples towards the longitudinal performance of iris recognition in children over 6.5 years from 13 sessions. Performance analysis is based on longitudinal match score (MS) distribution, FNMR, and  Linear Mixed Effects Regression Model to understand the impact of TF, EA, probe dilation, and $\Delta D$ on the MS. In addition to this, an in-depth analysis is performed to understand the root causes of the false rejection cases.  
\subsection{Longitudinal Performance Analysis: Match Score}
 VeriEye renders an MS of 1580 for a
perfect match and 0 for a non-match. In this study, a false non-match rate (FNMR) has been computed at the MS threshold of 36, corresponding to 0.1\% FMR. Fig~\ref{fig:MSDist} shows a graphical representation of MS distribution for 13 TFs between 6 months and 78 months (6.5 years) spaced by approximately 6 months. In two TFs (54 and 60) the number of subjects are too few to make reliable conclusions due to missed collections because of COVID. The match scores show a widespread distribution between 100 and 700. This indicates the possible impact of quality factors, like dilation, delta dilation, common usable iris area between the mating samples, etc. However, there is no clear trend of a decaying distribution across TFs. A detailed analysis is performed by modelling the MS as a factor of multiple parameters in Section~\ref{sec:LMER}. The blue line represents MS of 36. False rejections (FR), i.e., MS below 36, are depicted with red dots in the figure. Each dot in the figure represents an average MS per subject overall comparisons for that TF. A more detailed discussion on the cases of FR would be found in Section~\ref{sec:FalseRejections}.

\begin{figure}[h]
\centering

    \includegraphics[width=8.5cm]{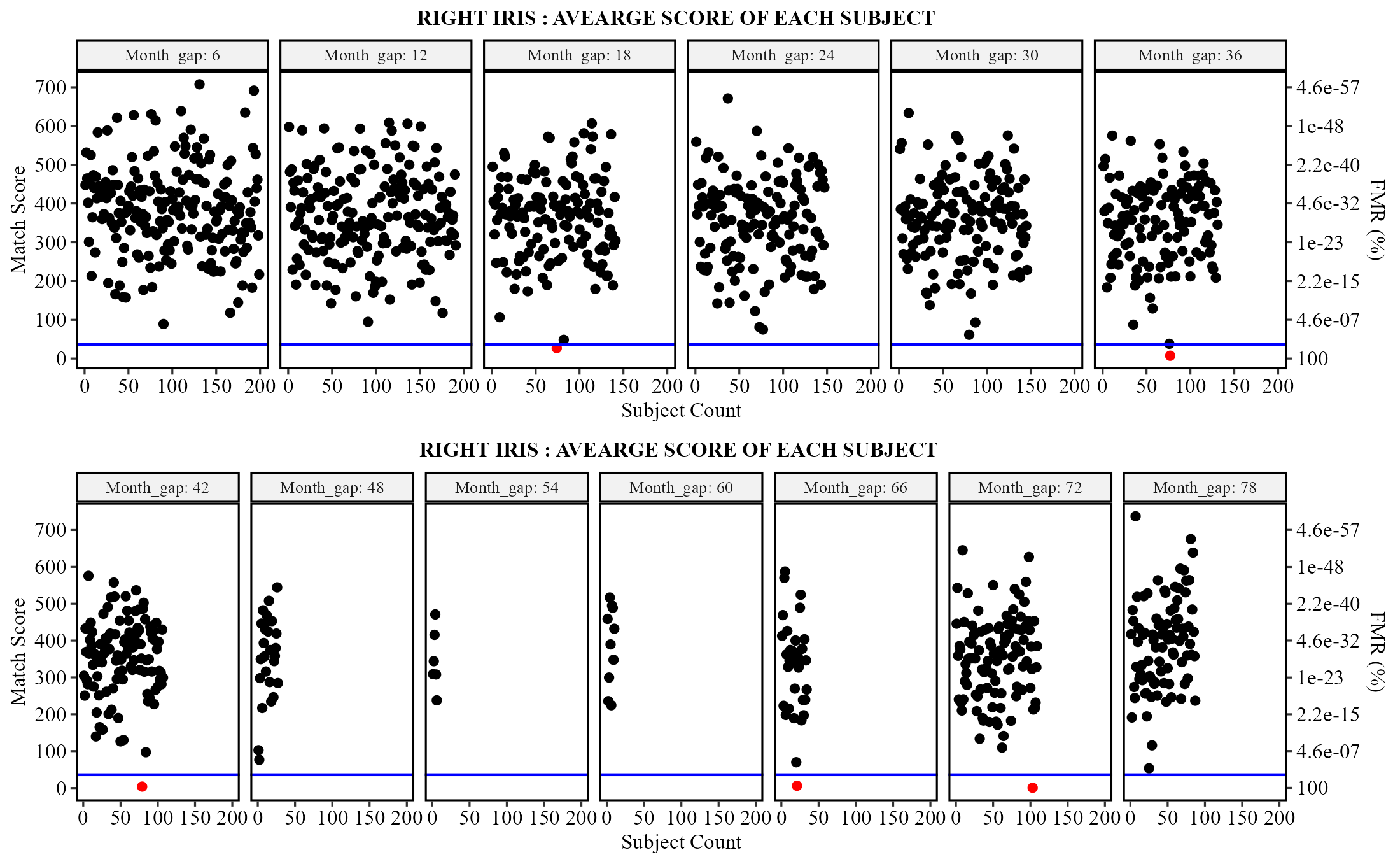}
    \caption{\footnotesize Match score distribution across 13 TFs of analysis across 6.5 years from the right iris. Each dot represents the average MS per subject per TF. The blue line represents the operating threshold of 36; red dots represent false rejections }
    \label{fig:MSDist}
    \vspace{-3mm}
 \end{figure}
Of the 30535 comparisons performed on iris samples from 230 subjects over 13 TFs, six instances of false rejection occurred from 4 subjects at fragmented TFs. FNMR over 13 TFs for both left and right iris is graphically presented in Fig~\ref{fig:FNMR}. FNMR of our dataset ranges between 0 to 0.80 \% at different TFs. No rejection was noted at 12, 24, 42, 48, 54, 60, 66 and 78-month TF. For TFs where no errors are noted, we adapted the `Rule of 3' for projected FNMR following ISO/IEC 19795-1\cite{iso2006iec}. `Rule of 3' computed the rate of occurrence of a particular event in a population which does not occur in an experiment as 3/N, where N is the sample size. 

 \begin{figure}[!t]
\centering
    \includegraphics[width=3.4in, height= 1.75in]{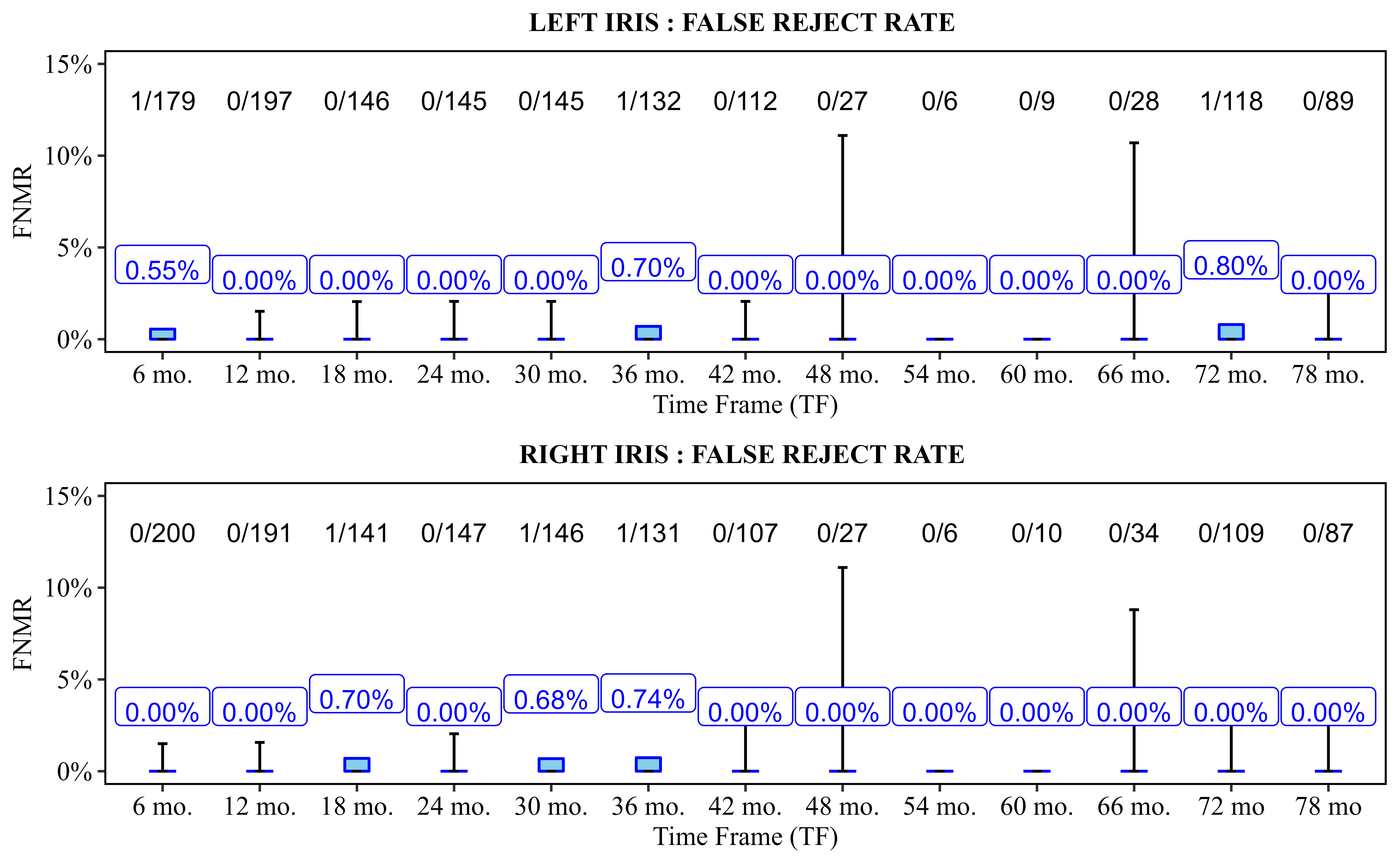}
    \caption{\footnotesize FNMR between 6 to 78 months TF from enrollment for left and right iris. Values marked in blue indicate actual FNMR at 0.1\% FMR. Error bars show projected FNMR based on the 'Rule of 3'. X/Y above each bar indicates the number of subjects rejected/total number of subjects at that TF. }
    \label{fig:FNMR}
    \vspace{-3mm}
 \end{figure}

\subsection{Longitudinal Performance Analysis: LMER}\label{sec:LMER}
Linear Mixed Effects Regression Model (LMER) is adapted to statistically understand \textit{the impact of explanatory variables - time difference (TD), enrollment age (EA), probe sample dilation (PD), and difference in dilation between mated pair of samples ($\Delta D$) on the response variable, the match score (MS).} The model takes into consideration \textit{random-effects} on the response variable by explanatory variables due to intra-subject variability, and \textit{fixed-effects} on the response variable by explanatory variables due to inter-subject variability. Equation~\ref{eq:Full Model} shows the model design to predict the impact of multiple variability factors on the MS. This design accounts for 92.24\% (89.5\% for LI) variance in the MS following the computation of the r-squared value. The unaccounted 7.77\% variance in MS could be due to miscellaneous factors like illumination, noise, and physiological condition, which were not considered as explanatory variables in the model.  

\begin{equation}
\scriptsize
\label{eq:Full Model}
    \begin{split}
    \mathit{\mathbf{MS \sim  \boldsymbol{\beta}_{0} + \boldsymbol{\beta}_{1}TF  +\boldsymbol{\beta}_{2} EA + \boldsymbol{\beta}_{3} PD + \boldsymbol{\beta}_{4} \Delta D +  \beta_{5} EA^2 + \beta_{6} \Delta PD^2 }}  \\
      \mathit{\mathbf{ + \boldsymbol{\beta}_{7} \Delta D^2  + b_{0i} +  b_{1i}TF  + b_{3i}PD + b_{4i} \Delta D + b_{6i} PD^2 }} 
    \end{split}
\end{equation}

 where,  
 \begin{itemize}[leftmargin=*,noitemsep]
  \item $\beta_{k}$ is the fixed regression coefficient for parameter, \textit{k}
   \item $b_{ki}$ is the random regression coefficient for subject, \textit{i}, parameter, \textit{k}
  
   \item $\beta_{0}+ b_{0i}$ is the sum of fixed and random intercepts corresponding to the initial state.
   
   \item $\beta_{k}+ b_{ki}$ is the subject specific (\textit{i}) gradient for the corresponding parameter \textit{k}.
   \item TF has been considered for both fixed and random effects to account for possible variability in MS due to both intra-subjects and inter-subject effects. 
   \item EA has been considered only for fixed effect. All other predictors are considered for both fixed and random effects. 
   \item Both linear and quadratic factors are introduced for variables: $\Delta D$, PD and EA for fixed effects. We assume the non-linearity in modelling MS for inter-subject data.
   
   \item Both linear and quadratic factors are considered for probe dilation in random effects for intra-subject data to account for any non-linearity in dilation for each subject over time.
   
   \item We consider the correlation between all coefficients associated with the same random effect term, i.e., for each subject, we consider the correlation between coefficients of TF, PD, $\Delta D$ and $ PD^2 $.
\end{itemize}
Detailed discussion on the selection and design of the model was presented in our previous report\cite{das2021iris}. Interpretation of the model results supports our research query's conclusions, as noted below. Results from LI and RI show a similar trend.
\begin{itemize}[noitemsep,leftmargin=*]
    \item TF, $\beta_{1}$, is negatively correlated with MS,  $\beta_{0}$ . However, the impact is both statistically and practically insignificant. 
    \item EA is negatively correlated with MS with statistical significance (p-value $<$ 0.001); the impact is linear; However, the estimated impact of EA, $\beta_{2}$, is practically insignificant with decay in MS of 4.02$\pm$0.73 with an increase in EA by one year. Thus, the MS estimate of 373.0 (LI) may decay by a maximum of 44.22 for EA of 11 years (maximum EA in our dataset). For an operating threshold of 36, the decay has no practical impact on performance. 
    \item Probe dilation shows the conflict in terms of statistical significance between RI and LI. In our modelling of the LI data, PD is statistically insignificant, whereas it is statistically significant for RI with p-value $<$0.05. In our dataset, the dilation of RI probe samples varies between 24 and 64. With the estimated decay of MS by 1.94 $\pm$ 0.93 due to $\beta_{3}$, for change is PD per unit, the decay in MS for our dataset may range between 46.56 $\pm$ 22.32 and 123.52 $\pm$ 59.52. 
    \item $\Delta D$, $\beta_{4}$, is negatively correlated with MS, $\beta_{0}$ with statistical and practical significance. The impact is non-linear. The $\Delta D$ in our dataset varies between 0 and 0.47. The maximum estimated decay in MS due to $\Delta D$ is 171.7 $\pm$ 26.37 with p-value $<$ 0.001. For an MS estimate, $\beta_{0}$, of 373.0 $\pm$ 8.72, an approximate decay of 197 is substantial; however, with the operating threshold being 36, the decay will not impact iris recognition performance. 
    \item Ordering covariates based on adding random variability (intra-subject) to the MS following Table~\ref{table:RE}: \\  $\Delta D$ $>$ Subject $>$ PD $>$ TF 
\end{itemize}
We conclude, \textbf{no impact of aging on MS across 6.5 years longitudinally} and the most sensitive factor impacting MS is $\Delta$D.

\begin{table}[!t]
\scriptsize
    \centering
    \caption{\textbf{Fixed Effects for Left and Right Iris}}
    \begin{tabular}{|c|c|c|c|}
    \hline
        \textbf{Variable} & \textbf{Parameter} &  \vtop{\hbox{\strut \textbf{Left Iris}}\hbox{\strut (\textbf{Est $ \pm $ SE)}}}  & \vtop{\hbox{\strut \textbf{Right Iris}}\hbox{\strut \textbf{(Est $ \pm $ SE)}}} \\
        \hline
        \hline
         Intercept & $\beta_{0}$ &  375.88 $\pm$ 8.60***  &  376.1 $\pm$ 7.99*** \\
         \hline
         TD & $\beta_{1}$ & -0.09 $\pm$ 0.11 NS  & - 0.17 $\pm$ 0.11 NS\\
         \hline
         EA & $\beta_{2}$ & -4.80 $\pm$ 0.78***  & -2.55 $\pm$ 0.83**  \\
         \hline
         PD & $\beta_{3}$ & -0.90 $\pm$ 0.75 NS  & -1.94 $\pm$ 0.93* \\
         \hline
        $\Delta D$ & $\beta_{4}$ &  -366.8 $\pm$ 55.74*** & -314.2 $\pm$ 66.11*** \\
        \hline
        $EA^2$ & $\beta_{5}$ & -0.73 $\pm$ 0.38 NS & -0.263 $\pm$ 0.38 NS\\
        \hline
        $PD^2$ & $\beta_{6}$ & -0.17 $\pm$ 0.12 NS & -0.02 $\pm$ 0.13 NS \\
        \hline
        $\Delta D^2 $ & $\beta_{7}$ &  -1253.56 $\pm$ 271.50***  & -994.4 $\pm$ 296.9***\\
        \hline
        \multicolumn{4}{c}{Significance Code: 0 `***' 0.001 `**' 0.01 `*' 0.05 `.' 0.1 ` ' 1 ; *** indicates}\\
        \multicolumn{4}{c}{ p-value between 0 and 0.001 with significance level 0.001 and so on.}\\
        \multicolumn{4}{c}{Est.: Estimate, SE: Standard Error, NS: Not significant}\\

    \end{tabular}
   
    \label{tab:fixed_effect}
\end{table}

\begin{table}[!t]
\scriptsize
\centering
\caption{\textbf{Random Effects Left and Right  Iris}}
\begin{tabular}{|*{4}{c|}}
\hline
\multirow{2}{*}{\textbf{Groups}}  &  \multirow{2}{*}{\textbf{Parameter}} & \multicolumn{2}{c|}{\textbf{Standard Deviation}} \\
 \cline{3-4}
&  & Left Iris & Right Iris \\
\hline
\hline
 Intercept & $b_{0i}$  & 121.08 & 111.09 \\
\hline
 TD & $b_{1i}$  & 1.49  & 1.49 \\
\hline
PD & $b_{3i}$ & 8.72 & 11.26 \\
\hline
$\Delta D$ & $b_{4i}$ & 647.24 & 827.29 \\
\hline
$PD^2$ & $b_{6i}$ & 1.46 &  1.70\\
\hline
Residual &   & 50.75 & 49.24 \\
\hline
\end{tabular}
\label{table:RE}
\vspace{-3mm}
\end{table}

\subsection{Longitudinal Performance Analysis: False Rejections}\label{sec:FalseRejections}
We performed a detailed assessment of the root causes of the 4 instances of recognition failure. Noting our overall observations on the failures - (a) None of the subjects were rejected on both the left and right iris; (b) Only a portion of the total genuine comparisons performed per subject at a particular TF was rejected. Multiple samples are collected per session per subject leading to multiple genuine comparisons for each TF. Rejection of a few samples while acceptance of other samples from the same session indicates an anomaly in the sample collected rather than a permanent change in the iris structure due to aging leading to failure. (c) Visual assessment of the samples revealed different poor quality factors of data leading to failure.

A summarization of the rejections of four subjects is tabulated in Table~\ref{table:FR}. Two of four subjects who were falsely rejected, subject ID 20160104994 and 20181012017, were analyzed in our earlier report\cite{das2021iris}. Both cases of false rejections were attributed to `poor quality'- poor enrollment sample and high rotation in the angle of presentation of the iris to the camera. The other two subjects are analyzed in detail in this report. Fig~\ref{fig:FR} shows the cases of false rejections and their corresponding match score distribution across the 6.5 years time frame for the sessions the subjects participated. Additionally, we looked at the dilation constancy (DC), the similarity in pupil dilation in a mated pair of iris images(\ref{DC}), of the false rejection samples with the enrollment samples. 
\begin{equation}
\footnotesize
    \mathit{\mathbf{Dilation\:Constancy(DC) = \frac{100 - max(D1,D2)}{100 - min(D1,D2)}}}
    \label{DC}
\end{equation}  

\textbf{False Reject Case 1:} Subject 20160104024 was rejected in 2 of the 4 comparisons of the left iris performed at 72 months time frame. Visual assessment revealed occlusion of a partial portion of the lower part of the enrollment sample by the lower eyelid. One of the two probe samples that caused the failures shows occlusion of a large portion of the upper iris by the upper eyelid and the eyelashes. Thus, \textit{low common usable iris area} between the probe and the enrollment sample for matching, leading to a very low match score below the operating threshold of the matcher. 

\textbf{False Reject Case 2:} Subject 2016051816008 was rejected in 14 of the 30 comparisons in the left iris in the 36 months TF of the 6 TFs it participated. All rejections were related to a common probe sample with hair occlusions and reflections on the eyeglass. \textit{Poor sample quality} led to false rejection. 

Our assessment concluded different forms of \textit{poor quality samples} as the root causes of false rejections. Our analysis did not indicate changes in the iris structure in children leading to failures in iris recognition performance over a longitudinal time frame of 6.5 years. Thus, we conclude \textit{no impact on the tails of the distribution due to aging in children.} Though our research directs towards the sensitivity of DC or $\Delta D$ on recognition performance, none of the cases of the FR in this study were caused by dilation. The FR samples had high DC. 

 \begin{figure}[!t]
\centering
    \includegraphics[width=3.4in]{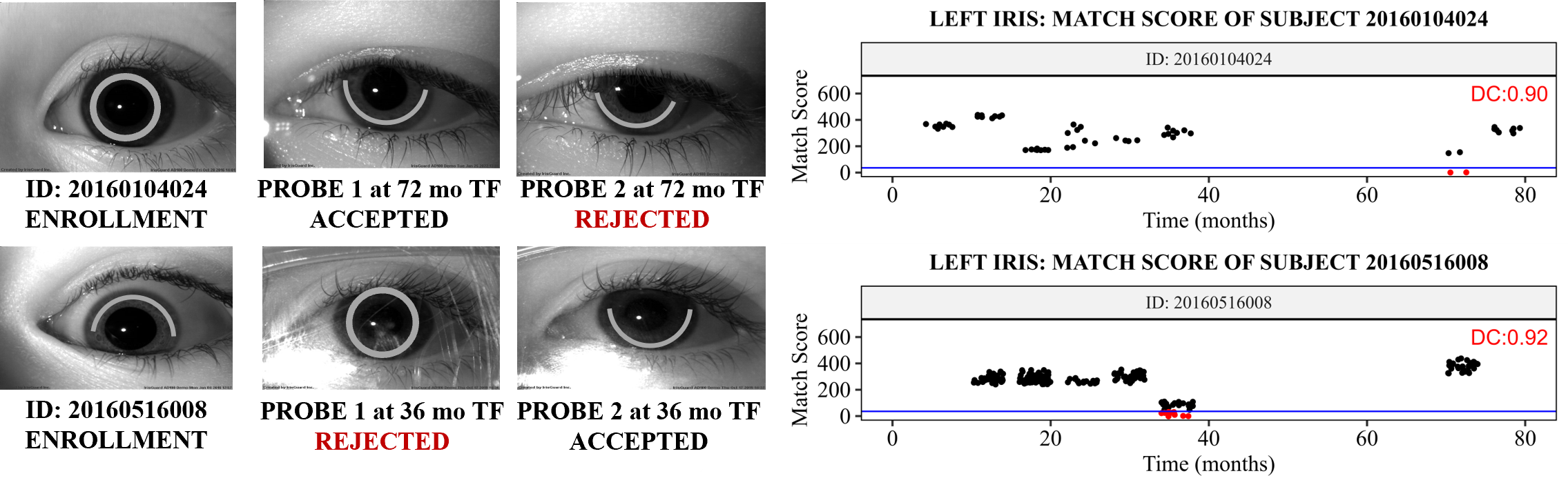}
    \caption{\footnotesize False rejection cases with their MS distribution longitudinally and average dilation constancy(DC); Blue line is the operational threshold at 36; red dots are false rejections. Note: Circles/semi-circles are drawn on the iris to obscure the iris patterns to preserve identity }
    \label{fig:FR}
    \vspace{-3mm}
 \end{figure}

\begin{table}[!ht]
\scriptsize
\centering
\caption{\textbf{False Rejection Case Summary across 6.5 years}}
\label{table:FR}
\begin{tabular}{|c|c|c|c|c|}
\hline
 \textbf{Coded ID} &  \textbf{Rejection TF} &  \textbf{Rejections} &  \textbf{Nature of Failure} &  \textbf{Eye} \\
 
 \hline
 20160104024 & 72  & 2 of 4 comp. &  Low Common Iris Area  & LI \\
 \hline
     20160104994 & 18, 30, 36 & 22 of 96 comp. & Poor Enrollment & RI\\
    \hline
  20160516008 & 36  & 14 of 30 comp. &Hair occlusion;Eye Glass & LI \\
  \hline
   20181012017 & 6  & 4 of 8 comp. & Angle of Presentation & LI\\
   \hline

\end{tabular}
\vspace{-3mm}
\end{table}

\section{Discussion and Conclusion}
\textit{Does aging change the iris structure to a point that impacts iris recognition performance for the demography of children?} For a complete understanding of the impact of aging and to make conclusions on this existing research question, a longitudinal dataset from the same children from birth to 18 years is essential. Such a dataset is not available for research. Possible causes could be the sensitivity of the data posing a challenge for data collection,  the Institutional Review Board (IRB) regulations for
children\cite{IRB_protocol} and the longitudinal commitment needed to pursue this research query. A systematic prospective longitudinal study is being conducted by our research team to scientifically understand the impact of aging in children on iris recognition performance. Our project started with children aged 4 to 11 years at enrollment and has created a longitudinal dataset from the same 230 children over 6.5 years. The study is ongoing. Though our dataset does not span across all ages for enrollment, the effective ages present in the dataset span between 4 years and 17.5 years. 6.5 years makes up one-third of the longitudinal time span for the children's demography. This is the largest study performed on longitudinal iris recognition performance in children.

Earlier, our research group published two reports on the findings from the analysis of the longitudinal iris data for 1-year \cite{johnson2018longitudinal} and 3 years \cite{das2021iris} as the research progressed. This report provides an update on the performance of iris recognition in children across 6.5 years longitudinally. 11223 samples from 230 subjects are analyzed and the report presented is based on 30535 mated comparisons of iris samples. The right and left iris were analyzed separately. False Non-Match Rate is adapted as a performance metric and cases of non-matches are analyzed in depth to understand the root cause. All of the six instances of false rejections from four subjects were attributed to different poor-quality samples. No instance of false rejection due to aging impact on the biometrics has been noted. Additionally, we assessed the impact of multiple variability factors (enrollment age, probe image dilation, difference in dilation between mated pair of samples and time difference between capture of enrollment and probe sample) by adopting a Linear Mixed Effects Regression Model (LMER).

Beyond the analysis presented in this paper, there still remains scope for additional studies. Is there a relationship between age, aging and dilation? If there is a relationship, does it impact recognition performance? Is there performance variation between younger and older age groups? We would report on these in the future. The minimum age at which iris recognition becomes viable is beyond the scope of this project as ages below four are not included in our dataset.

Based on our analysis, quality is the most important factor to maintain iris recognition performance. Our data is captured under a controlled environment being conscious of the factors impacting image quality. State-of-the-art iris capture systems having an in-built feature for quality checks are used for data collection. Even so, multiple cases of poor-quality samples were captured, some even leading to false rejections. Thus we emphasize the importance of stringent quality checks during enrollment to ensure a high-quality dataset eliminating recognition errors.

Our research conclusions are limited and cannot be extended beyond the ages and the timeframe studied. We conclude by summarizing the most important observations and conclusions from this analysis: (a) \textit{No impact on iris recognition performance (FNMR) due to aging} for ages 4 to 11 years over 6.5 years timeframe if the quality of the data is maintained. (b) The impact of increased time between enrollment and probe on MS is \textit{statistically and practically insignificant}. Analysis of additional data over 3.5 years deviated from our earlier conclusion \cite{das2021iris} where we noted statistically significant but practically insignificant decay in MS with time. (c) Delta dilation is the most impactful quality factor on MS variance; thus controlling the dilation is essential at the time of capture.

\section*{Acknowledgement}
We extend our gratitude to the Potsdam Elementary, Middle, and High School administration, staff, students, and the parents of the participants who have supported our research and the greater goal of scientific contribution to society. This project
would not have been successful without the efforts and 
valuable time put in by all data collectors. This work is financially supported by
Center for Identification Technology Research (CITeR) and National Science Foundation (NSF)(\# 1650503).

\bibliographystyle{IEEEtran}
\bibliography{bibliography}

\end{document}